\pdfoutput=1

\documentclass[11pt]{article}

\usepackage{ACL2023}

\usepackage{times}
\usepackage{latexsym}

\usepackage{bbding}
\usepackage{amsmath}
\usepackage{amssymb}
\usepackage{arydshln}
\usepackage{bbm} 
\usepackage{booktabs}
\usepackage{CJKutf8}
\usepackage{combelow}
\usepackage{enumitem}
\usepackage{epsfig}
\usepackage{fancybox}
\usepackage{float}
\usepackage{graphicx}
\usepackage{hyperref}
\usepackage{multirow}
\usepackage{multicol}
\usepackage{pgfplots}
\usepackage{pifont}
\usepackage{setspace}
\usepackage{subfigure}
\usepackage{tabularx}
\usepackage{url}
\usepackage{xspace}
\usepackage[vlined,ruled,commentsnumbered,linesnumbered]{algorithm2e}

\usepackage[T1]{fontenc}

\usepackage[utf8]{inputenc}

\usepackage{microtype}
\usepackage{enumitem}

\usepackage{inconsolata}

\title{Controllable Data Augmentation for Few-Shot Text Mining with\\ Chain-of-Thought Attribute Manipulation}

\author{Letian Peng \and Yuwei Zhang \and Jingbo Shang\thanks{$\ $  Corresponding author. } \\
University of California, San Diego \\
  \texttt{\{lepeng, yuz163, jshang\}@ucsd.edu}
  }

\newcommand{\method}{CoTAM\xspace}
\newcommand{\smallsection}[1]{\noindent\textbf{#1}.}
\newcommand{\fullmethod}{Chain-of-Thought Attribute Manipulation\xspace}

\begin{document}
\maketitle
\begin{abstract}
Prompting large language models (LLMs) for data augmentation has recently become a common practice in few-shot NLP tasks.
In this paper, we propose \fullmethod (\method), a novel approach that generates new data from existing examples by only tweaking in the user-provided, task-specific attribute, e.g., sentiment polarity or topic in movie reviews. 
Instead of conventional latent representation controlling, 
we leverage the chain-of-thought prompting to directly edit the text in three steps, (1) attribute decomposition, (2) manipulation proposal, and (3) sentence reconstruction.
Extensive results on various tasks, such as text (pair) classification, aspect-based sentiment analysis, and conditional text generation,
verify the superiority of \method over other LLM-based augmentation methods with the same number of training examples for both fine-tuning and in-context learning. 
Remarkably, the 2D visualization of the augmented dataset using principal component analysis revealed a human-recognizable decision boundary that is likely hinted by the attribute manipulation, demonstrating the potential of our proposed approach.

\end{abstract}

\section{Introduction}

\begin{figure}[t]
    \centering
    \includegraphics[width=0.49\textwidth]{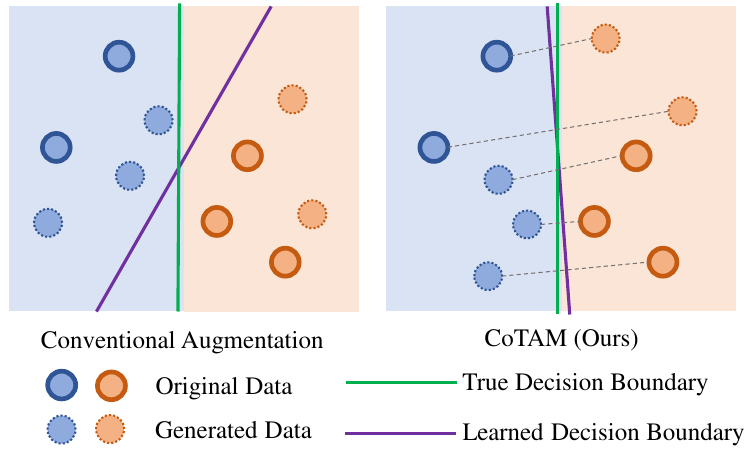}
    \caption{
    An illustrative comparison in case of binary classification. Conventional data augmentation generates uncontrolled data, while CoTAM directly reflects decision boundaries through task instructions. We present a real example in Figure~\ref{fig:pca}.}
    \vspace{-3mm}
    \label{fig:intro}
\end{figure}

Prompting large language models (LLMs) for data augmentation has recently become a common practice in few-shot natural language processing (NLP) tasks.
Existing methods~\cite{yoo2021gpt3mix,sahu2022data,auggpt,lin-etal-2023-selective} typically first generate new task-specific data with LLMs hinted by few-shot demonstrations and then fine-tune a (small) pre-trained language model with the augmented dataset for better performance. 
The same augmented data can be also incorporated into in-context learning (ICL) \cite{li2023dail,dong2023icl_survey}.
However, these augmentation methods usually prompt LLMs to generate new examples \emph{wildly} without proper control, which hinders the informativeness of generated data and might induce spurious correlation. 
As shown in Figure~\ref{fig:intro} (left), the generated data without control has no clear pattern and could even possibly mislead the fine-tuning or ICL under few-shot supervision.

In this paper, we propose a controllable data augmentation for few-shot text mining. 
The general idea is to generate new data from existing examples by only tweaking the user-provided, task-specific attribute, e.g., sentiment polarity or topic in movie reviews. 
Intuitively, as shown in Figure~\ref{fig:intro}, one can expect that this approach can efficiently find the decision boundary because we (1) directly manipulate along the direction of task-specific attributes and (2) maintain the rest of the attributes as before.

Different from the existing controllable generation works in computer vision~\cite{gan_manipulation, latent_semantics} and natural language processing~\cite{kruengkrai-2019-learning,flipda}, where attributes are manipulated in the latent space of the encoder before reconstructing new instances, we leverage the chain-of-thought (CoT) prompting~\cite{cot} to directly edit the text using LLMs in three steps, (1) attribute decomposition, (2) manipulation proposal, and (3) sentence reconstruction.
Specifically, 
we start with the user-provided, task-specific attributes, and then prompt LLMs to decompose each individual text example into other orthogonal attributes.
Compared with a pre-defined attribute set per dataset, we believe that such dynamically constructed, per-example sets of attributes can better capture the uniqueness of every piece of text.
Second, we instruct LLMs to propose a plan to manipulate the values of the task-specific attributes while maintaining the other attribute values the same.
Finally, we prompt the LLMs to reconstruct the sentence based on the manipulation proposal. 
All these steps are written in a single prompt and fed to the LLM at once.
Furthermore, using LLMs benefits the interpretability of our framework where attributes are completely transparent to users.

We conduct extensive experiments to evaluate \method and baselines using a series of few-shot classification tasks with very different classification targets, aspect-based sentiment analysis, and conditional text generation for more complex attribute manipulation.
For a fair comparison, all compared methods utilize the same LLMs and generate the same amount of data. 
We assess the quality of generated data by looking at (1) the performance of trained small language models via fine-tuning or tuning-free methods on the augmented data
and (2) the ICL performance of LLMs using the augmented data as demonstrations.
Extensive experimental results including label-scarce and out-of-domain scenarios demonstrate the advantage of proposed controllable data augmentation over conventional methods. The ablation study further reveals the necessity of attribute manipulation comparing to directly flipping the labels. Finally, we present PCA analysis on the embeddings of generated augmentations that visually illustrates the effectiveness of method.

Our contributions are three-fold:
\begin{itemize}[nosep,leftmargin=*]
    \item We propose a novel controllable data augmentation approach \method based on chain-of-thoughts prompting using LLMs, which directly edits the text examples in an interpretable way instead of tweak latent representation vectors. 
    \item We conduct experiments on a wide spectrum of tasks and datasets, demonstrating the effectiveness of the augmented data by \method in both fine-tuning and in-context learning.
    \item Our detailed analyses, especially the human-recognizable decision boundaries revealed by the 2D visualization of the augmented dataset using principle component analysis, demonstrate the significant potential of our proposed attribute manipulation approach. 
\end{itemize}

\smallsection{Reproducibility} We will open-source the code. \footnote{Code: \url{https://github.com/anonymous_repo}}

\begin{figure*}[t]
    \centering
    \includegraphics[width=0.99\linewidth]{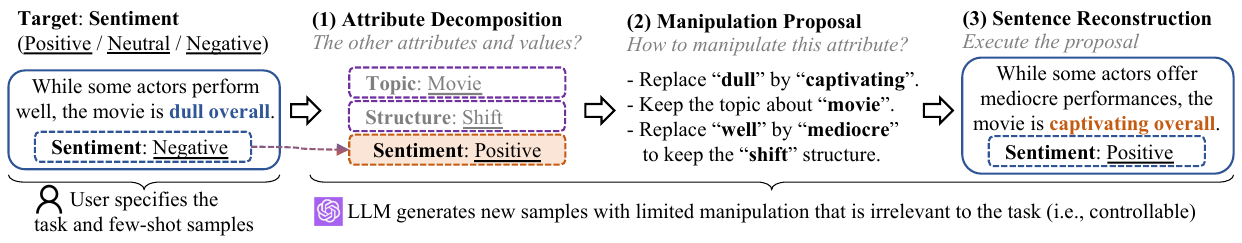}
    \caption{An overview of the goal and implementation of our \method.}
    \label{fig:overview}
\end{figure*}

\section{Problem Formulation}

We aim to generate more efficient training data using controllable augmentation on a few-shot dataset $\mathcal{D}$ focusing on a \textbf{target attribute} $Y$ (e.g., the classification objective) with $N$ possible values $\{y_1, y_2, \cdots, y_N\}$ (i.e., \textbf{$N$-way}). 
For each possible attribute value $y_i$, the dataset $\mathcal{D}$ provides $K$ examples (i.e., \textbf{$K$-shot}) of texts with the value. 
We here showcase two mainstream few-shot learning schemes as the basis to discuss the augmentation:
\begin{itemize}[nosep,leftmargin=*]
\item \textbf{In-context Learning} (ICL) is a scheme for LLMs, which takes a few examples of sentences with their target attribute values (i.e., a series of $(X, y_i)$) as the context to handle new inputs. 
With these demonstrations, the LLM is expected to understand the underlying mapping and then predict the label of new inputs. 
\item \textbf{Fine-tuning} generally trains smaller models with the (limited) labeled data. 
The model has a text embedder $\mathbf{E}$ and a classifier $\mathbf{C}$. 
A text $x$ from the dataset $\mathcal{D}$ will be represented as a dense vector $\mathbf{E}(x)$, which is learned to encode the attributes of $x$, including the target attribute $Y$ and other attributes. 
The classifier $\mathbf{C}$ further processes the vector $\mathbf{E}(x)$ and outputs a distribution over $y_1, y_2, \cdots, y_N$, indicating the probability of each $Y$ value in $x$. 
\end{itemize}
Ideally, our controllable augmentation shall supply efficient demonstrations and training data under the ICL and fine-tuning settings, respectively.

\section{Our \method Framework}


To boost the performance of few-shot methods, we suppose a scenario, shown in Figure~\ref{fig:overview}, to augment examples that well improve the task awareness of the inference models. For a given sample $x$ with target attribute value $y$ from $\mathcal{D}$, we will manipulate its attribute value to $y'$ that $y\neq y'$ to form a build a new sentence $x'$. 
We set two requirements for the manipulation: 1) Significant Manipulation on the target attribute $Y$, which means the manipulated result $x'$ should be viewed with $y_j$ by oracle like humans. 2) Minor Manipulation on all other attributes $\mathcal{Z}$, which indicates $x$ and $x'$ to share a similar value $z_k$ for all $Z \in \mathcal{Z}$. 
To meet the two requirements above will ensure $x$ and $x'$ only differ in attribute $Y$, making them an efficient pair for learning on the dataset $\mathcal{D}$. Take fine-tuning as an example, the loss $\mathcal{L}(X, y_i)+\mathcal{L}(X', y_j)$ will be attributed to the only different attribute $Y$, thus let each annotation by humans efficiently reflect the target attribute with its augmentations. 

Based on our desiderata above, we propose \method that benefits from the strong text manipulation capability of LLMs~\cite{chatgpt} with its workflow demonstrated in Figure~\ref{fig:overview}. 
To be more specific, we first create chain-of-thought (CoT) queries to decompose the input texts into many attributes, which approximates the latent space. 
We aim to get rid of human labor to propose other possible attributes for efficiency. Moreover, in some cases, even human experts cannot give you a complete list of other attributes among all the possible texts.
Finding a shared and fixed set of attributes for various kinds of texts is hard since different sentences rarely share a common set of applicable attributes. 
Encouraged by ~\citet{goal_cluster}, we instruct LLMs to propose a dynamic attribute set for each input text, which are customized among inputs dependent on which attributes are applicable.
The CoT then switches the value of the target attribute to other possible values in the task and prompts the LLM to reconstruct the manipulated sentence.
Finally, the LLM is guided to compose such a sentence to finalize the manipulation. 

Different from the existing controllable generation works in computer vision~\cite{gan_manipulation, latent_semantics} and natural language processing~\cite{kruengkrai-2019-learning,flipda}, where attributes are manipulated in the latent space of the encoder before reconstructing new instances, our \method is proposed to directly edit the text using LLMs.



\subsection{Step 1: Attribute Decomposition} Following the macro-level design of CoTAM, the first step in the CoT is to decompose the sentence into various attributes. The LLM takes the sentence and a human-annotated attribute-value pair as the input and then propose other attributes and their values.

For example, The sentence ``\textit{While some actors perform well, the movie is dull overall}'' with be processed with its known attribute-value $y_i$, here is ``\textit{Sentiment: Negative}''. 
The LLM then proposes a set of other applicable attribute-values $\mathcal{\hat{Z}} = \textrm{LLM}_{\textrm{AD}}(X, y_i) \subset \mathcal{Z}$ like ``\textit{Topic: Moive}'', ``\textit{Structure: Shift}'' as in Figure~\ref{fig:overview}, which is a subset of $\mathcal{Z}$ but is generally detailed enough to approximate the irrelevant attributes. 
The value of the known attribute is then flipped to another one given by the user like ``\textit{Sentiment: Positive}'', which is then combined with other LLM-proposed attribute-values for the next step. 


\subsection{Step 2: Manipulation Proposal} 

In the second step, we will instruct the LLM to propose the methodology to reconstruct a sentence with the switched attribute and others from the decomposition step. 
This step is incorporated as understanding how to achieve the goal, which is important to the CoT inference~\cite{cot}. 
In this step, the LLM takes all elements in the manipulation as the input and proposes an instruction $I = \textrm{LLM}_{\textrm{MP}}(X, y_i, y_j, \mathcal{\hat{Z}})$ for LLM to execute in the next step. A proposed manipulation is shown as in Figure~\ref{fig:overview}, the LLM suggest several instructions to complete the manipulation. 

\subsection{Step 3: Sentence Reconstruction}

This step simply asks the LLM to follow its proposed manipulation instruction $I$ to reconstruct the sentence and output a label-flipped one as $X' = \textrm{LLM}_{\textrm{SR}}(X, I)$. As in Figure~\ref{fig:overview}, the LLM follows the self-generated instructions to edit the input sentence to generate our desired $X'$ that has significant different in $Y$ (sentiment polarity) and minor difference in $\hat{Z}$ (proposed other attributes).

\section{Experiments}

In this section, we evaluate different LLM-based augmentation methods on a series of classification tasks, with different target attributes. 
We incorporate comprehensive ways of utilizing augmentations with different classification techniques, such as fine-tuning, in-context learning and inference with sentence embedding. 
We further evaluate the augmentation ability of methods on more complex tasks like aspect-based sentiment analysis and conditional text generation.

\subsection{Datasets}

\begin{table}
\centering
\small
\scalebox{.75}{
\begin{tabular}{lll}
\toprule
\textbf{Dataset} & \textbf{Target Attribute} & \textbf{Possible Value} \\
\midrule
SST-2 & Sentiment & Positive \\ 
TweetEmo & Sentiment & Anger \\ 
AG-News & Topic & World News \\ 
MNLI & Natural Language Inference & Contradiction \\ 
MRPC & Semantics & Equivalent to Sentence 1 \\ 
CSQA & Best choice & <Answer Name> \\ 
ABSA & Sentiment on <Aspect> & Positive \\ 
CommonGen & Keywords & "ski", "mountain", "skier" \\ 
\bottomrule
\end{tabular}
}
\caption{Target attributes and possible values in datasets of our experiments and more details can be found in Appendix~\ref{apdx:attr}.} 
\label{tab:attr_inst}
\end{table}

We verify the advantage of \method on text classification and other tasks using $6$ classification datasets, including \textbf{SST-2} (sentiment polarity) \cite{sst2}, \textbf{TweetEmo} (fine-grained sentiment) \cite{tweeteval}, AG-NEWS (topic) \cite{ag}, \textbf{MNLI} (natural language inference) \cite{mnli}, \textbf{MRPC} (semantic textual similarity) \cite{mrpc}, and \textbf{CSQA} (multiple choice question answering) \cite{csqa}. MNLI includes matched (\textbf{MNLI$_{\textrm{m}}$}) and mismatched (\textbf{MNLI$_{\textrm{mm}}$}) datasets for evaluation. To further test the ability of \method on attributes other than classification targets, we include a manipulation on aspect-based sentiment analysis (ABSA) and conditional text generation tasks. For ABSA datasets, we include \textbf{Restaurant} and \textbf{Laptop} from SemEval2014 \cite{ABSA14}. For conditional text generation, we include \textbf{CommonGen} \cite{commongen}. We report the results on $1000$ samples for ICL from the mixture of validation and test datasets due to cost issues. For other setups, we report results on the validation dataset when the test dataset is not publicly available considering the efficiency to get multi-run results. The statistics of datasets are presented in Appendix~\ref{apdx:stat}. We present some examples of attribute names in Table~\ref{tab:attr_inst}.

\subsection{Compared Methods}

\begin{table*}
\centering
\small
\begin{tabular}{llccccccc}
\toprule
\multicolumn{2}{l}{\textbf{Method}} & SST-2 & TweetEmo & AG-NEWS & MNLI$_{\textrm{m}}$ & MNLI$_{\textrm{mm}}$ & MRPC & CSQA \\
\midrule
\multirow{6}*{\rotatebox{90}{Fine-tuning}} & Base & $60.54$ & $44.38$ & $81.05$ & $35.88$ & $38.75$ & $51.96$ & $34.54$ \\
& Extra Annotation$^\dag$ & $62.17$ & $69.51$ & $88.66$ & $43.33$ & $44.03$ & $57.50$ & $47.36$ \\
& LLM Pseudo Label & $61.14$ & $69.11$ & $85.64$ & $41.71$ & $42.92$ & $55.88$ & $45.12$ \\
& FlipDA++ & $74.28$ & $70.87$ & $84.72$ & $51.52$ & $53.56$ & $60.15$ & $50.52$ \\
& CoTDA & $70.83$ & $67.76$ & $85.19$ & $36.06$ & $36.28$ & $55.54$ & $48.79$ \\
& \method & $\textbf{79.12}$ & $\textbf{72.76}$ & $\textbf{85.80}$ & $\textbf{54.07}$ & $\textbf{56.16}$ & $\textbf{65.83}$ & $\textbf{53.22}$ \\
\midrule
\multirow{7}*{\rotatebox{90}{ICL}} & No Example & $90.50$ & $69.80$ & $81.30$ & $67.50$ & $\textbf{69.70}$ & $69.80$ & $73.50$ \\
& Base & $94.00$ & $74.50$ & $85.50$ & $68.10$ & $68.10$ & $70.60$ & $76.30$ \\
& Extra Annotation$^\dag$ & $94.70$ & $79.00$ & $88.70$ & $68.70$ & $68.60$ & $71.40$ & $76.80$ \\
& LLM Pseudo Label & $94.20$ & $75.80$ & $85.80$ & $66.90$ & $69.00$ & $67.90$ & $76.50$ \\
& FlipDA++ & $94.30$ & $76.70$ & $85.20$ & $68.80$ & $68.90$ & $70.70$ & $77.00$ \\
& CoTDA  & $94.00$ & $76.50$ & $86.00$ & $68.20$ & $68.50$ & $70.00$ & $76.70$ \\
& \method & $\textbf{94.50}$ & $\textbf{77.10}$ & $\textbf{86.40}$ & $\textbf{69.70}$ & $69.20$ & $\textbf{70.90}$ & $\textbf{77.30}$ \\
\bottomrule
\end{tabular}
\caption{Few-shot learning results based on data annotated by humans and LLMs. $\dag$: Extra Annotation increases the number (NK) of human-annotated samples to the same number as LLM-annotated to compare the annotation ability between LLMs and humans. \textbf{Bold:} The best result with the base number (K) of human annotation, thus excluding ``Extra Annotation''.}
\label{tab:main}
\end{table*}


\paragraph{CoT Data Augmentation (CoTDA)} is an augmentation variant of our method that applies a similar CoT for conventional augmentation. Instead of directly asking for augmentation, we let the LLM follow our proposed CoT and propose a methodology to write a sentence with the \textbf{same} attributes as the input sentence. CoTDA is the main baseline for comparison to explore the importance of attribute switching in our CoTAM. For each seed data, we augment it for N-1 times with $0.1$ temperature, where N refers to the number of classes in the dataset. Thus, CoTDA generates the same number of new data as CoTAM to achieve a fair comparison.

\paragraph{FlipDA} \cite{flipda} is a traditional label-switched augmentation method based on conditional generation by a fully-tuned T5 \cite{T5}. Specifically, the sentence is combined with the switched label as the input to T5. Then, some spans in the sentence are randomly masked and recovered by T5 conditioning on the new label to switch the semantics of the sentence. As the original FlipDA requires a large supervised dataset that is inapplicable to few-shot learning, we build an LLM-based FlipDA (\textbf{FlipDA++}) baseline by sending span replacement instructions to LLMs. 

\paragraph{Human/LLM Annotation} directly using the texts labeled by humans or LLMs. For human annotation, we include the K-shot (\textbf{Base}) and NK-shot (\textbf{Extra Annotation}) setups. K-shot represents the baseline before integrating the data generated from LLMs. NK-shot has the number of training data after augmentation with human annotation, thus we expect it to be a upper bound of augmentation methods. Whereas, we will see CoTAM able to outperform this upper bound, which can be attributed to higher data quality resulting from attribute manipulation. NK-shot LLM annotation\footnote{K-shot data are used for in-context inference.} (\textbf{Pseudo Label}) represents a simple baseline that is generally applied when much unlabeled in-domain data is available.

\paragraph{Comparison Fairness} We select \texttt{GPT-4} \cite{chatgpt} as the LLM to construct the dataset. The temperature of GPT-4 to set to $0$ towards high quality and reproducibility. 
We apply each augmentation method to a fixed subset of each dataset to create a small subset from which we sample training data. 
For a fair comparison, this subset is also used in other baselines for data generation. By default, we set K to $10$ for fine-tuning and $3$ to ICL. All reported results are the average over $10$ runs (except for ICL due to expense) to eliminate the bias.

All the prompts in our experiments are presented in Appendix~\ref{apdx:prompt} for better reproducibility.\footnote{To further increase reproducibility, we also include results on open-sourced LLMs in Appendix~\ref{apdx:open}}

\subsection{Classification Result}

\begin{table}
\centering
\small
\resizebox{\linewidth}{!}{
\begin{tabular}{lcccccc}
\toprule
\multirow{2}*{\textbf{Method}} & \multicolumn{2}{c}{SST-2} & \multicolumn{2}{c}{TweetEmo} & \multicolumn{2}{c}{AG-NEWS} \\
\cmidrule(l){2-3} \cmidrule(l){4-5} \cmidrule(l){6-7} 
 & NC & KNN & NC & KNN & NC & KNN \\
\midrule
Base & $82.00$ & $78.20$ & $66.01$ & $59.92$ & $77.72$ & $73.57$ \\
Extra$^\dag$ & $87.55$ & $83.45$ & $71.23$ & $67.56$ & $84.70$ & $82.33$ \\
LLM SL & $86.78$ & $80.26$ & $69.34$ & $64.90$ & $\textbf{81.19}$ & $\textbf{79.34}$ \\
FlipDA++ & $88.13$ & $86.76$ & $66.53$ & $65.05$ & $79.82$ & $75.11$ \\
CoTDA & $86.38$ & $83.00$ & $68.63$ & $61.58$ & $78.87$ & $76.56$ \\
\method & $\textbf{88.43}$ & $\textbf{87.52}$ & $\textbf{70.02}$ & $\textbf{65.37}$ & $80.60$ & $75.48$ \\
\bottomrule
\end{tabular}
}
\caption{Utilization of sentence embeddings for classification tasks based on different augmented few-shot examples.}
\label{tab:instance}
\end{table}

\paragraph{Fine-tuning} A simple way to evaluate the data quality is to tune a model on it and then check its performance. We select \texttt{RoBERTa-Large} \cite{roberta} as the learner on different datasets. With the validation dataset unavailable, we train the model for $32$ epochs\footnote{Except $8$ epochs for MRPC, on which the model is more likely to overfit.} and then evaluate it. 

As presented in Table~\ref{tab:main}, our \method achieves the best fine-tuning results on all $7$ tasks in comparison with other LLM-based data generation methods. On most tasks, the two label-switching methods (FlipDA and \method) outperform other methods, which indicates using the LLM to switch labels creates more efficient data. On label switching, attribute manipulation shows superiority over simple span replacement as our \method performs better than FlipDA on all tasks. The prominent performance of \method also verifies the capability of LLMs to manipulate complex attributes which might refer to premises or questions. 

On $6$ out of $7$ tasks, our \method breaks the supposed upper boundary of (N-way) NK-shot with extra human annotations. This indicates that carefully crafted data from LLMs have the potential to train better models than ones trained on the same number of human annotations. Also, aur \method is verified to be such a method that improves the data efficiency by attribute manipulation. 

\paragraph{In-context Learning} The performances of ICL-based inference with different augmentation methods are demonstrated in Table~\ref{tab:main}. Our \method show superior ability on providing LLMs with few-shot examples for inference, thus broadening the application of our method. The only fail case for \method is the out-of-domain MNLI, where few-shot examples do not benefit the inference. Still, among all augmentation scenarios, our \method performs the best for this evaluation. 

\paragraph{Inference w/ Sentence Embedding} In the field of few-shot text classification, text embedding has proven to be a powerful tool for improving performance and efficiency \cite{mteb}. This section is dedicated to exploring instance-based techniques designed explicitly for text classification with text embedding models.

In instance-based inference, a text embedding model converts the input sentence into a representation. The label of this representation is then determined based on its proximity to annotated sentence representations. We utilized two tuning-free algorithms in our experiments—Nearest Centroid (NC) \cite{nearest_centroid} and K-Nearest Neighbors (KNN)—and applied them to three different text classification datasets.
NC assigns a label to an input sentence depending on how close it is to centroids, defined as the average representation of sentences sharing the same label. In contrast, KNN labels the input sentence according to the most common label amongst its nearest K neighbors. We set K to 5 in our experiments.
We harness the Simple Contrastive Sentence Embedding (SimCSE) model \cite{simcse}, with \texttt{RoBERTa-Large} as the backbone model\footnote{\href{https://huggingface.co/princeton-nlp/sup-simcse-roberta-large}{huggingface.co/princeton-nlp/sup-simcse-roberta-large}}, to encode the texts.

Table~\ref{tab:instance} showcases the performance of different data generation methods when used with instance-based algorithms. In contrast to methods that generate new texts (such as FlipDA and CoTDA), our proposed method, referred to as \method hereafter, exhibits superior performance in most configurations. This implies that data created by \method also benefits from improved distributions in the latent space of text embedding models.
On the AG-NEWS dataset, instance-based algorithms show a preference for in-domain annotations, whether made by humans or Large Language Models (LLMs). This highlights the importance of using in-domain texts when employing these algorithms for certain tasks. 

\subsection{Aspect-based Sentiment Analysis}

\begin{table}[t]
\centering
\small
\resizebox{\linewidth}{!}{
\begin{tabular}{lcccccc}
\toprule
\multirow{2}*{\textbf{Method}} & \multicolumn{3}{c}{Restaurant} & \multicolumn{3}{c}{Laptop} \\
\cmidrule(l){2-4} \cmidrule(l){5-7} 
 & P. & R. & F. & P. & R. & F. \\
\midrule
Base & $30.61$ & $40.38$ & $34.82$ & $23.73$ & $28.57$ & $25.93$ \\
Extra$^\dag$ & $54.70$ & $66.67$ & $60.09$ & $59.18$ & $44.62$ & $50.88$ \\
LLM SL & $44.26$ & $56.25$ & $49.54$ & $18.56$ & $22.73$ & $14.09$ \\
FlipDA++ & $45.90$ & $58.33$ & $51.38$ & $26.58$ & $42.86$ & $32.81$ \\
CoTDA & $44.55$ & $51.04$ & $47.57$ & $26.09$ & $36.74$ & $30.51$ \\
CoTAM & $\textbf{50.00}$ & $\textbf{64.58}$ & $\textbf{56.36}$ & $\textbf{33.33}$ & $\textbf{44.90}$ & $\textbf{38.26}$ \\
\bottomrule
\end{tabular}
}
\caption{The performance of span manipulation on aspect-based sentiment analysis datasets.}
\label{tab:absa}
\end{table}

Here we further expand the utility of CoTAM to a more complex scenario to manipulate multiple span representations. We experiment on aspect-based sentiment analysis (ABSA), which aims to extract spans targeted by sentiment (aspects) in a statement and corresponding polarities. For instance, the aspect extracted from ``The food is good.'' will be ``positive aspect: food''.

For attribute manipulation on ABSA, we view the aspects as the ABSA attributes like ``positive aspect: food''. We query the LLMs to decompose texts into ABSA and other attributes. The polarities of ABSA attributes are then randomly switched and used for the reconstruction. The reconstructed data are merged into the initial dataset as the augmentation.

We use the SemEval$2014$ ABSA dataset which has two subsets: restaurant and laptop and three sentiment polarities: positive, negative, and neutral\footnote{We remove the conflict polarity because of its sparsity in the dataset.}. We set the shot number ($K$) to $10$ and generate $2$ times for each instance ($N=3$), which is the maximal manipulation time for an instance with only one aspect. The results on ABSA are presented in Table~\ref{tab:absa}, our \method successfully outperforms other LLM-based augmentation methods, which confirms that \method is applicable to more complex scenarios than single sentence attribute manipulation.

\begin{table}[t]
\centering
\small
\scalebox{0.95}{
\begin{tabular}{lcccccc}
\toprule
\multirow{2}*{\textbf{Method}} & \multicolumn{4}{c}{CommonGen} \\
\cmidrule(l){2-5}
 & Rouge-1 & Rouge-2 & Rouge-L & Coverage \\
\midrule
Base & $41.99$ & $13.98$ & $33.57$ & $65.07$ \\
Extra$^\dag$ & $46.30$ & $15.66$ & $36.18$ & $75.55$ \\
LLM SL & $\textbf{47.85}$ & $14.68$ & $35.63$ & $75.95$ \\
FlipDA++ & $46.81$ & $14.48$ & $35.32$ & $\textbf{76.10}$ \\
CoTDA & $44.43$ & $13.43$ & $35.05$ & $65.98$ \\
CoTAM & $46.38$ & $\textbf{15.85}$ & $\textbf{37.02}$ & $75.23$ \\
\bottomrule
\end{tabular}
}
\caption{The performance of span manipulation on conditional text generation. \textbf{Coverage} means the ratio of given keywords that appeared in the output sentence.}
\label{tab:commongen}
\end{table}

\subsection{Conditional Text Generation}

We run experiments on the CommonGen dataset to apply our \method to conditional text generation. CommonGen targets to generate a sentence that contains a set of keywords. For instance, with input words: ``ski, mountain, skier'', the output can be ``Skier skis down the mountain'' We use CoTAM to manipulate the data by switching the group of keywords to another one (proposed by the LLM) while keeping other attributes unchanged.

According to the metrics shown in Table~\ref{tab:commongen}, we can view CoTAM holds its advantage over other LLM-based augmentation methods. Thus, we conclude that CoTAM is not only limited to classification tasks but also has the potential for information extraction and natural language generation.

\section{Further Analysis}

\subsection{Workflow Demonstration}

\begin{figure}
\centering
\includegraphics[width=0.49\textwidth]{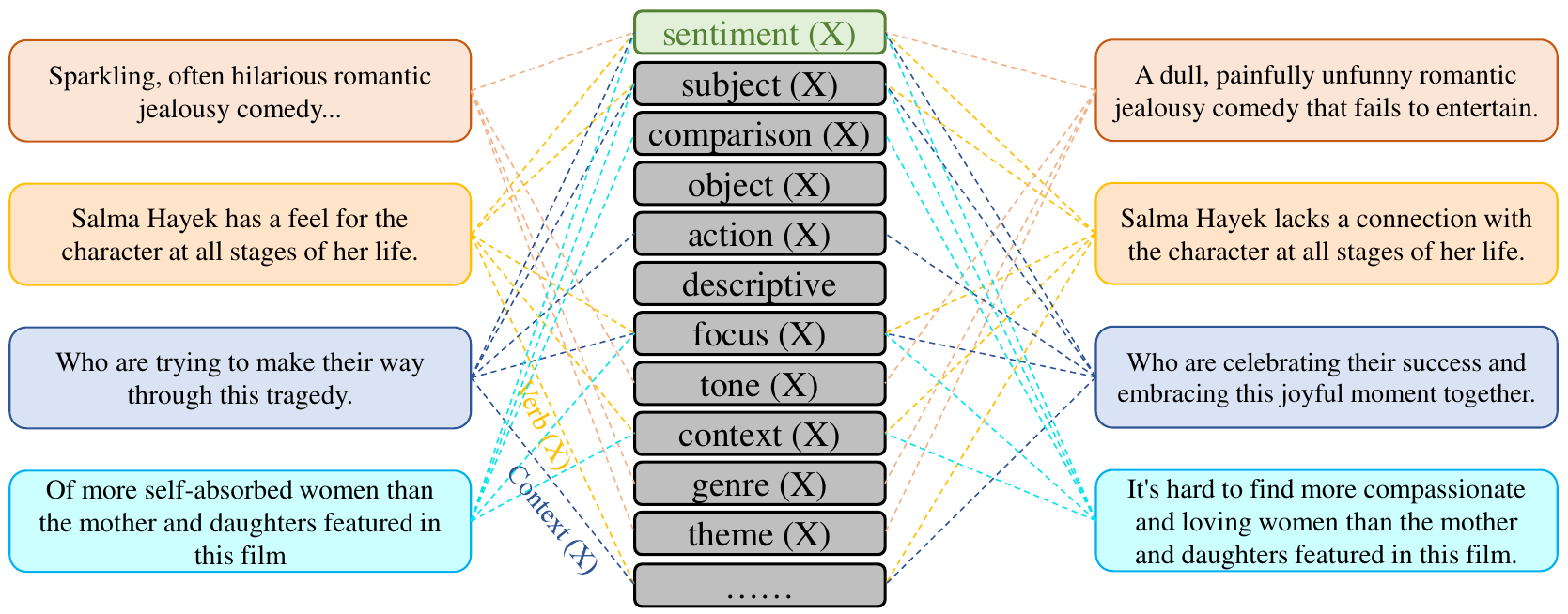}
\caption{The workflows of CoTAM for different inputs.}
\label{fig:workflow}
\end{figure}

In Figure~\ref{fig:workflow}, we demonstrate the workflow of the dynamic attribute decomposition mechanism. We include the attributes that most commonly appear in the manipulation according to the statistics in Appendix~\ref{apdx:stats}. In the workflow, our CoTAM decomposes sentences into applicable attributes and reconstructs while maintaining these attributes. For instance, \textit{tone (X)} is more applicable to the first sentence due to its subjectivity and \textit{comparison (X)} is more applicable to the last sentence since only it involves comparison. These attributes comprehend the unchanged parts of texts to guide the reconstruction during the manipulation. Subsequently, the reconstruction switch the targeted label (\textit{sentiment (X)} in the case) with minor change to other attributes. 

\subsection{Ablation Study}

\begin{table}
\centering
\small
\begin{tabular}{lcccc}
\toprule
\multirow{2}*{\textbf{Data}} & \multicolumn{3}{c}{SST-2} & MNLI \\
\cmidrule(l){2-4} \cmidrule(l){5-5} 
& T & NC & KNN & T \\
\midrule
CoTAM & $\textbf{79.12}$ & $\textbf{88.43}$ & $\textbf{87.52}$ & $\textbf{54.07}$ \\
\quad w/o What & $75.69$ & $88.03$ & $86.78$ & $45.61$ \\
\quad w/o How & $77.94$ & $88.15$ & $87.01$ & $48.98$ \\
\quad w/o CoT & $71.82$ & $87.94$ & $86.24$ & $39.34$ \\
\quad w/ V$3.5$ & $72.93$ & $87.59$ & $84.31$ & $41.32$ \\
\quad w/ FAP & $76.38$ & $87.79$ & $85.13$ & $47.91$ \\
\bottomrule
\end{tabular}
\caption{The ablation study on our CoTAM. Matched MNLI results are presented for analysis.}
\label{tab:ablation}
\end{table}

We launch an ablation study to verify the importance of each thought in the CoT. 
We also explore the effect of different LLMs. We thus change the LLM in our experiments to \texttt{GPT-3.5-turbo}. The experiments show that the \texttt{GPT-4} leads to significantly better fine-tuning results. Also, this gap can be narrowed down by text embedding models on text classification.

The outcomes of our ablation study are detailed in Table~\ref{tab:ablation}. In this study, we found that eliminating each ``thought'' from our CoT resulted in a decline in performance. Interestingly, the ``what'' (decomposition) thought proved more critical than the ``how'' (methodology) thought, accentuating the predominance of attribute proposal over auxiliary methodology proposal. The CoT is necessary for label switching as the removal of it leads to significant performance degradation. In comparison between LLMs, \texttt{GPT-4} outperforms \texttt{GPT-3.5-turbo}, indicating that \method favors larger LLM with better language capability, especially on more complex tasks like MNLI. Finally, we compare the performance of between CoTAM with a fixed attribute pool (FAP) and with a dynamic attribute pool in our experiments. The result shows the advantage to remove the type limitation of attribute the LLM decomposes into.

\subsection{Visualization of Attribute Manipulation}

\begin{figure}
\centering
\includegraphics[width=0.49\textwidth]{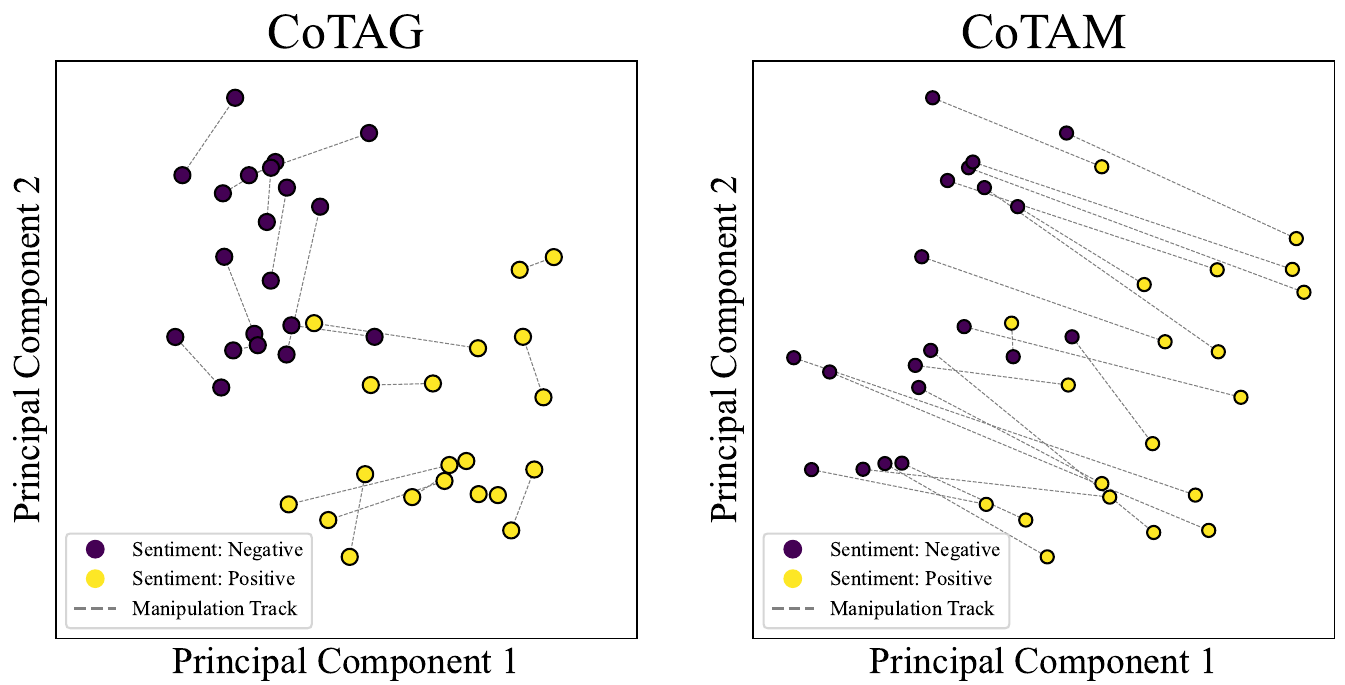}
\caption{Principal component analysis of text pairs generated by our CoTDA and \method on the SST-2 dataset.}
\label{fig:pca}
\end{figure}

In an attempt to confirm our hypothesis that LLM is adjusting a single feature while keeping other attributes constant, we illustrate data pair representations from CoTAM in Figure~\ref{fig:pca}. We use principal component analysis (PCA) \cite{pca} to take the high-dimensional (1024-dimensional) text representations from SimCSE and simplify them into a 2-dimensional space for ease of visualization.

The diagram distinctly demarcates between positive and negative representations, which underscores the value of our method in fine-tuning and instance-based inference. Additionally, the direction of representation switching is largely consistent, providing further evidence that LLMs have the ability to tweak one attribute while keeping others stable. This consistency in the direction of the switch hints at the predictability and control we have exercised over LLM behavior for targeted feature manipulation. In comparison to CoTDA, our CoTAM depicts a clearer boundary, thus enabling more efficient data learning than traditional data augmentation.

\subsection{Data Scale Analysis}

\begin{figure}
\centering
\includegraphics[width=0.4\textwidth]{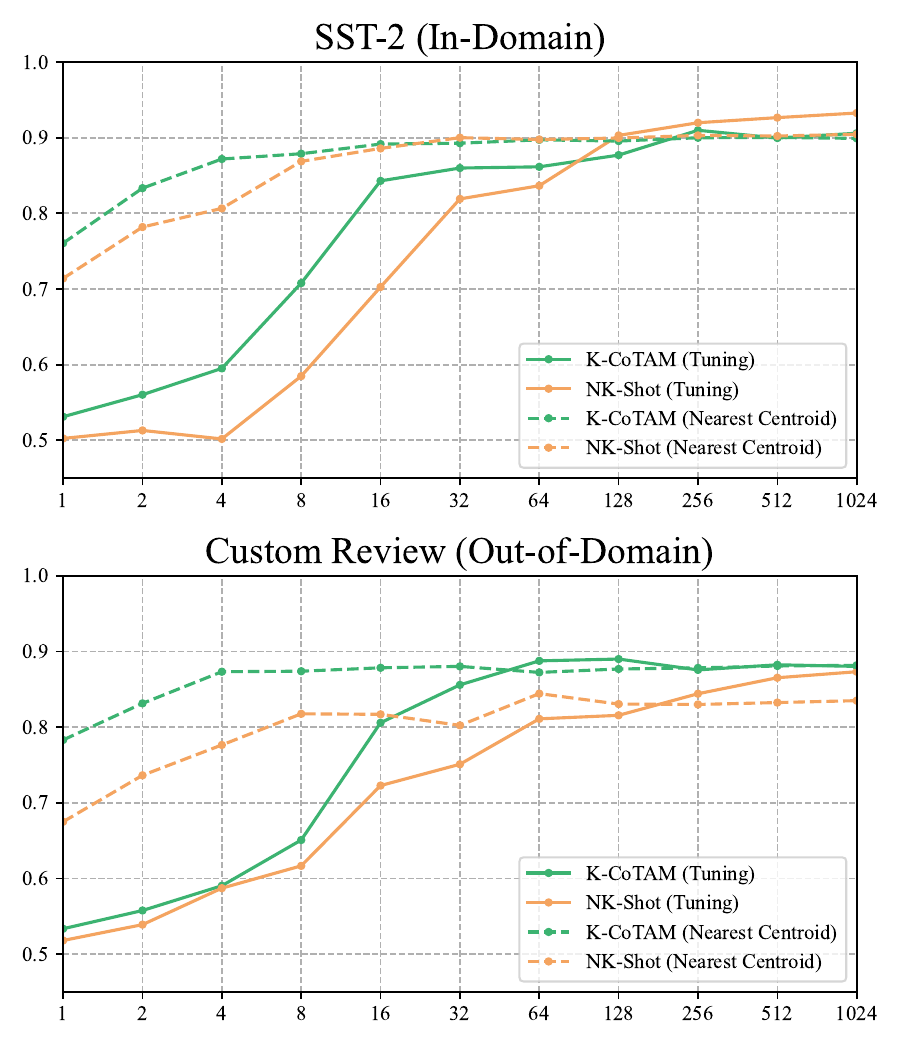}
\caption{Comparison between K-shot CoTAM and NK-shot on in-domain and out-of-domain test datasets.}
\label{fig:scale}
\end{figure}

In this section, we analyze how the number of initial data affects the performance of our \method. Thus, we sample $3000$ more instances from SST-2 to scale up the sampling pool. As presented in Figure~\ref{fig:scale}, \method is able to break the NK-Shot boundary with few examples ($K \leq 64$) for fine-tuning. With text representation models, \method shows a significant advantage on very few examples ($K \leq 4$) and converges to a similar performance with human annotation. Though fine-tuning on more human annotation leads to higher performance than \method, the in-domain performance improvement might be a result of overfitting to the domain. Thus, we further evaluate \method and NK-Shot on custom review, an out-of-domain dataset with the same labels as SST-2. On custom review, \method shows a consistent advantage with different data numbers. Thus, we conclude our \method is more robust to domain mismatching than direct tuning.

\subsection{Case Study}
\label{apdx:case}
\begin{figure*}[t]
\centering
\includegraphics[width=0.99\textwidth]{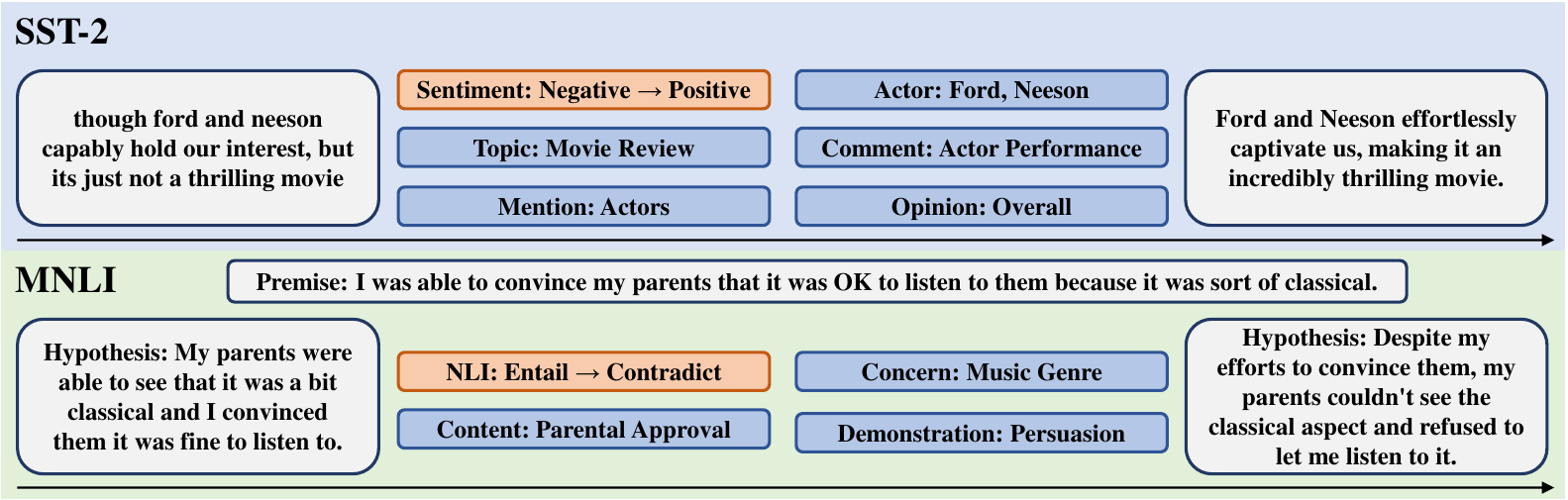}
\caption{Case study of the real workflow in \method.}
\label{fig:case}
\end{figure*}

Figure~\ref{fig:case} specifies the real attribute manipulation process in our experiments. For better depiction, we simplify the response by only presenting the attributes proposed by the LLMs. 

In the SST-2 example, other attributes include labels in a different categorization (Topic: Movie Review), actor entities (Actor: Ford, Neeson), and overall style (Opinion: Overall). These attributes are well preserved in the reconstruction, which contributes to a strong contrast in the task target and consequently improves the data efficiency.

Moving on to the MNLI example, the sentence primarily breaks down into different semantic elements. When these elements are reconstructed, they follow a logical sequence that differs from the original sentence. Thus data from \method reinforces the learner's comprehension of textual logic which is crucial for tackling MNLI.

\section{Related Work}



\noindent\textbf{Attribute Manipulation} aims to control certain attributes of the data. A general application of attribute manipulation is to change the visual attributes in facial images \cite{gan_manipulation, latent_semantics}. Image manipulation generally involves the transformation of image representations \cite{cgan,elegant,gan_manipulation} in the latent space. In natural language processing, the closest topic to attribute manipulation is data flipping \cite{sent_flip,flipda}, which replaces key spans in the text to switch its label. Obviously, many textual attributes like topics cannot be manipulated by span replacement. Thus, we choose to adapt the LLM to manipulate a latent space approximated by a series of attributes proposed by the LLM. 

\noindent\textbf{Controllable Generation} is another close topic to our CoTAM. These methods typically generate texts from a continuous latent space discretely by controlling certain dimensions \cite{generate_from_space,control_text,fudge}. The controllable generator is trained by maximizing a variational lower bound on the data log-likelihood under the generative model with a KL divergence loss \cite{control_text}. The limitation of the current controllable generation is no explicit control of other dimensions to maintain them the same. Our method addresses this issue by completely decomposing the input text into multiple labels with LLMs and then reconstructing it with switched attributes. 

\noindent\textbf{Large Language Models} are large-scale models trained on a massive number of texts~\cite{brown2020language,chowdhery2022palm,Hoffmann2022TrainingCL} that have been shown to have emerging capabilities~\cite{wei2022emergent}. One of these capabilities is learning from few-shot demonstrations, which is often referred to as in-context learning~\cite{dong2022survey}. However, these demonstrations must be concatenated into contexts during inference time, increasing the computational costs and carbon footprints. Another important capability is to follow instructions for zero-shot task transferrability~\cite{wei2022finetuned}. Following this idea, ChatGPT~\cite{ouyang2022training,chatgpt} was trained with human feedback and reinforcement learning. Our work benefits from these instruction-tuned models to generate attributes and reconstruct sentences.

\noindent\textbf{Data Augmentation}  is widely employed in low-resource scenarios to mitigate model overfitting. It is usually conducted in a label-preserving manner where only minor perturbations are added~\cite{wei-zou-2019-eda,fadaee-etal-2017-data}. Recently, a line of research propose to use LLMs for data augmentation. Specifically, they use few-shot data as demonstrations and prompt LLMs to generate new data~\cite{yoo2021gpt3mix,sahu-etal-2022-data}. They claim that the LLM is able to mix few-shot data and synthesize similar ones. \citealp{lin-etal-2023-selective} further propose to use Pointwise V-information to filter unhelpful data from generations. Most recently \citealp{auggpt,whitehouse2023llm} propose to generate data using ChatGPT and GPT-4 and observe performance improvement. Finally \citealp{cheng2023improving} use GPT-3 generated data to improve sentence embedding via contrastive learning.
Our work aims at improving LLM-based data augmentation via attribute manipulation.

\section{Conclusion}

The study introduces a novel method, \fullmethod (\method), which uses manipulated data from Large Language Models (LLMs) for few-shot learning. Our \method creates label-switched data by modifying task-specific attributes and reconstructing new sentences. Our testing validated the effectiveness of \method over other LLM-based text generation techniques. The results also showcase the potential for LLM-guided learning with less supervision. 

Future work will aim to adapt the attribute manipulation technique for smaller language models, increasing its accessibility. This would reduce reliance on the resource-intensive processes inherent to large language models, improving efficiency.

\section*{Limitation}

Despite the significant advancements in few-shot learning and attribute manipulation reported in this paper, our proposed \method does come with certain limitations. Firstly, our approach leverages a chain-of-thoughts decomposition and reconstruction procedure which, while yielding improved data efficiency and model performance, tends to result in a decrease in the overall generation efficiency compared to traditional methods. This may affect the method's scalability, particularly in scenarios requiring rapid data generation. Secondly, the current implementation of \method is primarily confined to attribute-related tasks, limiting its scope of application. While this constraint is a direct result of our method's design focused on manipulating task-specific attributes, we acknowledge that extending \method's applicability to a broader set of tasks could significantly increase its utility. Our future work will thus aim to address this limitation. Lastly, it should be noted that the effectiveness of \method is fundamentally dependent on the abilities of the underlying Large Language Models. As a consequence, the limitations inherent in these LLMs, such as biases in their training data or limitations in their understanding of nuanced contexts, could potentially impact the performance of \method. It is thus crucial to continually improve and refine the LLMs used in our method to ensure the accuracy and robustness of the generated data.

\section*{Ethical Consideration}

Our work instructs large language models to generate efficient training data, which generally does not raise ethical concerns.

\bibliography{anthology,custom}

\begin{thebibliography}{45}
\expandafter\ifx\csname natexlab\endcsname\relax\def\natexlab#1{#1}\fi

\bibitem[{Barbieri et~al.(2020)Barbieri, Camacho{-}Collados, Anke, and
  Neves}]{tweeteval}
Francesco Barbieri, Jos{\'{e}} Camacho{-}Collados, Luis~Espinosa Anke, and
  Leonardo Neves. 2020.
\newblock \href {https://doi.org/10.18653/v1/2020.findings-emnlp.148}
  {Tweeteval: Unified benchmark and comparative evaluation for tweet
  classification}.
\newblock In \emph{Findings of the Association for Computational Linguistics:
  {EMNLP} 2020, Online Event, 16-20 November 2020}, volume {EMNLP} 2020 of
  \emph{Findings of {ACL}}, pages 1644--1650. Association for Computational
  Linguistics.

\bibitem[{Bowman et~al.(2016)Bowman, Vilnis, Vinyals, Dai, J{\'{o}}zefowicz,
  and Bengio}]{generate_from_space}
Samuel~R. Bowman, Luke Vilnis, Oriol Vinyals, Andrew~M. Dai, Rafal
  J{\'{o}}zefowicz, and Samy Bengio. 2016.
\newblock \href {https://doi.org/10.18653/v1/k16-1002} {Generating sentences
  from a continuous space}.
\newblock In \emph{Proceedings of the 20th {SIGNLL} Conference on Computational
  Natural Language Learning, CoNLL 2016, Berlin, Germany, August 11-12, 2016},
  pages 10--21. {ACL}.

\bibitem[{Brown et~al.(2020)Brown, Mann, Ryder, Subbiah, Kaplan, Dhariwal,
  Neelakantan, Shyam, Sastry, Askell et~al.}]{brown2020language}
Tom Brown, Benjamin Mann, Nick Ryder, Melanie Subbiah, Jared~D Kaplan, Prafulla
  Dhariwal, Arvind Neelakantan, Pranav Shyam, Girish Sastry, Amanda Askell,
  et~al. 2020.
\newblock Language models are few-shot learners.
\newblock \emph{Advances in neural information processing systems},
  33:1877--1901.

\bibitem[{Cheng et~al.(2023)Cheng, Yang, Sun, Li, and Qiu}]{cheng2023improving}
Qinyuan Cheng, Xiaogui Yang, Tianxiang Sun, Linyang Li, and Xipeng Qiu. 2023.
\newblock Improving contrastive learning of sentence embeddings from ai
  feedback.
\newblock \emph{arXiv preprint arXiv:2305.01918}.

\bibitem[{Chowdhery et~al.(2022)Chowdhery, Narang, Devlin, Bosma, Mishra,
  Roberts, Barham, Chung, Sutton, Gehrmann et~al.}]{chowdhery2022palm}
Aakanksha Chowdhery, Sharan Narang, Jacob Devlin, Maarten Bosma, Gaurav Mishra,
  Adam Roberts, Paul Barham, Hyung~Won Chung, Charles Sutton, Sebastian
  Gehrmann, et~al. 2022.
\newblock Palm: Scaling language modeling with pathways.
\newblock \emph{arXiv preprint arXiv:2204.02311}.

\bibitem[{Dai et~al.(2023)Dai, Liu, Liao, Huang, Wu, Zhao, Liu, Liu, Li, Zhu,
  Cai, Li, Liu, and Li}]{auggpt}
Haixing Dai, Zhengliang Liu, Wenxiong Liao, Xiaoke Huang, Zihao Wu, Lin Zhao,
  Wei Liu, Ninghao Liu, Sheng Li, Dajiang Zhu, Hongmin Cai, Quanzheng Li,
  Tianming Liu, and Xiang Li. 2023.
\newblock Chataug: Leveraging chatgpt for text data augmentation.

\bibitem[{Dolan and Brockett(2005)}]{mrpc}
William~B. Dolan and Chris Brockett. 2005.
\newblock \href {https://aclanthology.org/I05-5002/} {Automatically
  constructing a corpus of sentential paraphrases}.
\newblock In \emph{Proceedings of the Third International Workshop on
  Paraphrasing, IWP@IJCNLP 2005, Jeju Island, Korea, October 2005, 2005}. Asian
  Federation of Natural Language Processing.

\bibitem[{Dong et~al.(2023)Dong, Li, Dai, Zheng, Wu, Chang, Sun, Xu, Li, and
  Sui}]{dong2023icl_survey}
Qingxiu Dong, Lei Li, Damai Dai, Ce~Zheng, Zhiyong Wu, Baobao Chang, Xu~Sun,
  Jingjing Xu, Lei Li, and Zhifang Sui. 2023.
\newblock \href {http://arxiv.org/abs/2301.00234} {A survey on in-context
  learning}.

\bibitem[{Dong et~al.(2022)Dong, Li, Dai, Zheng, Wu, Chang, Sun, Xu, and
  Sui}]{dong2022survey}
Qingxiu Dong, Lei Li, Damai Dai, Ce~Zheng, Zhiyong Wu, Baobao Chang, Xu~Sun,
  Jingjing Xu, and Zhifang Sui. 2022.
\newblock A survey for in-context learning.
\newblock \emph{arXiv preprint arXiv:2301.00234}.

\bibitem[{Fadaee et~al.(2017)Fadaee, Bisazza, and Monz}]{fadaee-etal-2017-data}
Marzieh Fadaee, Arianna Bisazza, and Christof Monz. 2017.
\newblock \href {https://doi.org/10.18653/v1/P17-2090} {Data augmentation for
  low-resource neural machine translation}.
\newblock In \emph{Proceedings of the 55th Annual Meeting of the Association
  for Computational Linguistics (Volume 2: Short Papers)}, pages 567--573,
  Vancouver, Canada. Association for Computational Linguistics.

\bibitem[{F.R.S.(1901)}]{pca}
Karl~Pearson F.R.S. 1901.
\newblock \href {https://doi.org/10.1080/14786440109462720} {Liii. on lines and
  planes of closest fit to systems of points in space}.
\newblock \emph{The London, Edinburgh, and Dublin Philosophical Magazine and
  Journal of Science}, 2(11):559--572.

\bibitem[{Gao et~al.(2021)Gao, Yao, and Chen}]{simcse}
Tianyu Gao, Xingcheng Yao, and Danqi Chen. 2021.
\newblock \href {https://doi.org/10.18653/v1/2021.emnlp-main.552} {Simcse:
  Simple contrastive learning of sentence embeddings}.
\newblock In \emph{Proceedings of the 2021 Conference on Empirical Methods in
  Natural Language Processing, {EMNLP} 2021, Virtual Event / Punta Cana,
  Dominican Republic, 7-11 November, 2021}, pages 6894--6910. Association for
  Computational Linguistics.

\bibitem[{Hoffmann et~al.(2022)Hoffmann, Borgeaud, Mensch, Buchatskaya, Cai,
  Rutherford, de~Las~Casas, Hendricks, Welbl, Clark, Hennigan, Noland,
  Millican, van~den Driessche, Damoc, Guy, Osindero, Simonyan, Elsen, Rae,
  Vinyals, and Sifre}]{Hoffmann2022TrainingCL}
Jordan Hoffmann, Sebastian Borgeaud, Arthur Mensch, Elena Buchatskaya, Trevor
  Cai, Eliza Rutherford, Diego de~Las~Casas, Lisa~Anne Hendricks, Johannes
  Welbl, Aidan Clark, Tom Hennigan, Eric Noland, Katie Millican, George van~den
  Driessche, Bogdan Damoc, Aurelia Guy, Simon Osindero, Karen Simonyan, Erich
  Elsen, Jack~W. Rae, Oriol Vinyals, and L.~Sifre. 2022.
\newblock Training compute-optimal large language models.
\newblock \emph{ArXiv}, abs/2203.15556.

\bibitem[{Hu et~al.(2017)Hu, Yang, Liang, Salakhutdinov, and
  Xing}]{control_text}
Zhiting Hu, Zichao Yang, Xiaodan Liang, Ruslan Salakhutdinov, and Eric~P. Xing.
  2017.
\newblock \href {http://proceedings.mlr.press/v70/hu17e.html} {Toward
  controlled generation of text}.
\newblock In \emph{Proceedings of the 34th International Conference on Machine
  Learning, {ICML} 2017, Sydney, NSW, Australia, 6-11 August 2017}, volume~70
  of \emph{Proceedings of Machine Learning Research}, pages 1587--1596. {PMLR}.

\bibitem[{Kruengkrai(2019{\natexlab{a}})}]{kruengkrai-2019-learning}
Canasai Kruengkrai. 2019{\natexlab{a}}.
\newblock \href {https://doi.org/10.18653/v1/D19-1659} {Learning to flip the
  sentiment of reviews from non-parallel corpora}.
\newblock In \emph{Proceedings of the 2019 Conference on Empirical Methods in
  Natural Language Processing and the 9th International Joint Conference on
  Natural Language Processing (EMNLP-IJCNLP)}, pages 6311--6316, Hong Kong,
  China. Association for Computational Linguistics.

\bibitem[{Kruengkrai(2019{\natexlab{b}})}]{sent_flip}
Canasai Kruengkrai. 2019{\natexlab{b}}.
\newblock \href {https://doi.org/10.18653/v1/D19-1659} {Learning to flip the
  sentiment of reviews from non-parallel corpora}.
\newblock In \emph{Proceedings of the 2019 Conference on Empirical Methods in
  Natural Language Processing and the 9th International Joint Conference on
  Natural Language Processing, {EMNLP-IJCNLP} 2019, Hong Kong, China, November
  3-7, 2019}, pages 6310--6315. Association for Computational Linguistics.

\bibitem[{Li et~al.(2023)Li, Li, Mekala, Li, wang, Wang, Hogan, and
  Shang}]{li2023dail}
Dawei Li, Yaxuan Li, Dheeraj Mekala, Shuyao Li, Yulin wang, Xueqi Wang, William
  Hogan, and Jingbo Shang. 2023.
\newblock \href {http://arxiv.org/abs/2311.03319} {Dail: Data augmentation for
  in-context learning via self-paraphrase}.

\bibitem[{Lin et~al.(2020)Lin, Zhou, Shen, Zhou, Bhagavatula, Choi, and
  Ren}]{commongen}
Bill~Yuchen Lin, Wangchunshu Zhou, Ming Shen, Pei Zhou, Chandra Bhagavatula,
  Yejin Choi, and Xiang Ren. 2020.
\newblock \href {https://doi.org/10.18653/V1/2020.FINDINGS-EMNLP.165}
  {Commongen: {A} constrained text generation challenge for generative
  commonsense reasoning}.
\newblock In \emph{Findings of the Association for Computational Linguistics:
  {EMNLP} 2020, Online Event, 16-20 November 2020}, volume {EMNLP} 2020 of
  \emph{Findings of {ACL}}, pages 1823--1840. Association for Computational
  Linguistics.

\bibitem[{Lin et~al.(2023)Lin, Papangelis, Kim, Lee, Hazarika, Namazifar, Jin,
  Liu, and Hakkani-Tur}]{lin-etal-2023-selective}
Yen-Ting Lin, Alexandros Papangelis, Seokhwan Kim, Sungjin Lee, Devamanyu
  Hazarika, Mahdi Namazifar, Di~Jin, Yang Liu, and Dilek Hakkani-Tur. 2023.
\newblock \href {https://aclanthology.org/2023.eacl-main.107} {Selective
  in-context data augmentation for intent detection using pointwise
  {V}-information}.
\newblock In \emph{Proceedings of the 17th Conference of the European Chapter
  of the Association for Computational Linguistics}, pages 1463--1476,
  Dubrovnik, Croatia. Association for Computational Linguistics.

\bibitem[{Liu et~al.(2019)Liu, Ott, Goyal, Du, Joshi, Chen, Levy, Lewis,
  Zettlemoyer, and Stoyanov}]{roberta}
Yinhan Liu, Myle Ott, Naman Goyal, Jingfei Du, Mandar Joshi, Danqi Chen, Omer
  Levy, Mike Lewis, Luke Zettlemoyer, and Veselin Stoyanov. 2019.
\newblock \href {http://arxiv.org/abs/1907.11692} {Roberta: {A} robustly
  optimized {BERT} pretraining approach}.
\newblock \emph{CoRR}, abs/1907.11692.

\bibitem[{Manning et~al.(2008)Manning, Raghavan, and
  Schütze}]{nearest_centroid}
Christopher~D. Manning, Prabhakar Raghavan, and Hinrich Schütze. 2008.
\newblock \href
  {http://nlp.stanford.edu/IR-book/information-retrieval-book.html}
  {\emph{Introduction to Information Retrieval}}.
\newblock Cambridge University Press, Cambridge, UK.

\bibitem[{Muennighoff et~al.(2023)Muennighoff, Tazi, Magne, and Reimers}]{mteb}
Niklas Muennighoff, Nouamane Tazi, Lo{\"{\i}}c Magne, and Nils Reimers. 2023.
\newblock \href {https://aclanthology.org/2023.eacl-main.148} {{MTEB:} massive
  text embedding benchmark}.
\newblock In \emph{Proceedings of the 17th Conference of the European Chapter
  of the Association for Computational Linguistics, {EACL} 2023, Dubrovnik,
  Croatia, May 2-6, 2023}, pages 2006--2029. Association for Computational
  Linguistics.

\bibitem[{OpenAI(2023)}]{chatgpt}
OpenAI. 2023.
\newblock \href {https://doi.org/10.48550/arXiv.2303.08774} {{GPT-4} technical
  report}.
\newblock \emph{CoRR}, abs/2303.08774.

\bibitem[{Ouyang et~al.(2022)Ouyang, Wu, Jiang, Almeida, Wainwright, Mishkin,
  Zhang, Agarwal, Slama, Gray, Schulman, Hilton, Kelton, Miller, Simens,
  Askell, Welinder, Christiano, Leike, and Lowe}]{ouyang2022training}
Long Ouyang, Jeffrey Wu, Xu~Jiang, Diogo Almeida, Carroll Wainwright, Pamela
  Mishkin, Chong Zhang, Sandhini Agarwal, Katarina Slama, Alex Gray, John
  Schulman, Jacob Hilton, Fraser Kelton, Luke Miller, Maddie Simens, Amanda
  Askell, Peter Welinder, Paul Christiano, Jan Leike, and Ryan Lowe. 2022.
\newblock \href {https://openreview.net/forum?id=TG8KACxEON} {Training language
  models to follow instructions with human feedback}.
\newblock In \emph{Advances in Neural Information Processing Systems}.

\bibitem[{Perarnau et~al.(2016)Perarnau, van~de Weijer, Raducanu, and
  {\'{A}}lvarez}]{cgan}
Guim Perarnau, Joost van~de Weijer, Bogdan Raducanu, and Jos{\'{e}}~M.
  {\'{A}}lvarez. 2016.
\newblock \href {http://arxiv.org/abs/1611.06355} {Invertible conditional gans
  for image editing}.
\newblock \emph{CoRR}, abs/1611.06355.

\bibitem[{Pontiki et~al.(2014)Pontiki, Galanis, Pavlopoulos, Papageorgiou,
  Androutsopoulos, and Manandhar}]{ABSA14}
Maria Pontiki, Dimitris Galanis, John Pavlopoulos, Harris Papageorgiou, Ion
  Androutsopoulos, and Suresh Manandhar. 2014.
\newblock \href {https://doi.org/10.3115/V1/S14-2004} {Semeval-2014 task 4:
  Aspect based sentiment analysis}.
\newblock In \emph{Proceedings of the 8th International Workshop on Semantic
  Evaluation, SemEval@COLING 2014, Dublin, Ireland, August 23-24, 2014}, pages
  27--35. The Association for Computer Linguistics.

\bibitem[{Raffel et~al.(2020)Raffel, Shazeer, Roberts, Lee, Narang, Matena,
  Zhou, Li, and Liu}]{T5}
Colin Raffel, Noam Shazeer, Adam Roberts, Katherine Lee, Sharan Narang, Michael
  Matena, Yanqi Zhou, Wei Li, and Peter~J. Liu. 2020.
\newblock \href {http://jmlr.org/papers/v21/20-074.html} {Exploring the limits
  of transfer learning with a unified text-to-text transformer}.
\newblock \emph{J. Mach. Learn. Res.}, 21:140:1--140:67.

\bibitem[{Sahu et~al.(2022{\natexlab{a}})Sahu, Rodriguez, Laradji,
  Atighehchian, Vazquez, and Bahdanau}]{sahu-etal-2022-data}
Gaurav Sahu, Pau Rodriguez, Issam Laradji, Parmida Atighehchian, David Vazquez,
  and Dzmitry Bahdanau. 2022{\natexlab{a}}.
\newblock \href {https://doi.org/10.18653/v1/2022.nlp4convai-1.5} {Data
  augmentation for intent classification with off-the-shelf large language
  models}.
\newblock In \emph{Proceedings of the 4th Workshop on NLP for Conversational
  AI}, pages 47--57, Dublin, Ireland. Association for Computational
  Linguistics.

\bibitem[{Sahu et~al.(2022{\natexlab{b}})Sahu, Rodriguez, Laradji,
  Atighehchian, Vazquez, and Bahdanau}]{sahu2022data}
Gaurav Sahu, Pau Rodriguez, Issam~H Laradji, Parmida Atighehchian, David
  Vazquez, and Dzmitry Bahdanau. 2022{\natexlab{b}}.
\newblock Data augmentation for intent classification with off-the-shelf large
  language models.
\newblock \emph{arXiv preprint arXiv:2204.01959}.

\bibitem[{Shen et~al.(2020)Shen, Gu, Tang, and Zhou}]{gan_manipulation}
Yujun Shen, Jinjin Gu, Xiaoou Tang, and Bolei Zhou. 2020.
\newblock \href {https://doi.org/10.1109/CVPR42600.2020.00926} {Interpreting
  the latent space of gans for semantic face editing}.
\newblock In \emph{2020 {IEEE/CVF} Conference on Computer Vision and Pattern
  Recognition, {CVPR} 2020, Seattle, WA, USA, June 13-19, 2020}, pages
  9240--9249. Computer Vision Foundation / {IEEE}.

\bibitem[{Shen and Zhou(2021)}]{latent_semantics}
Yujun Shen and Bolei Zhou. 2021.
\newblock \href {https://doi.org/10.1109/CVPR46437.2021.00158} {Closed-form
  factorization of latent semantics in gans}.
\newblock In \emph{{IEEE} Conference on Computer Vision and Pattern
  Recognition, {CVPR} 2021, virtual, June 19-25, 2021}, pages 1532--1540.
  Computer Vision Foundation / {IEEE}.

\bibitem[{Socher et~al.(2013)Socher, Perelygin, Wu, Chuang, Manning, Ng, and
  Potts}]{sst2}
Richard Socher, Alex Perelygin, Jean Wu, Jason Chuang, Christopher~D. Manning,
  Andrew~Y. Ng, and Christopher Potts. 2013.
\newblock \href {https://aclanthology.org/D13-1170/} {Recursive deep models for
  semantic compositionality over a sentiment treebank}.
\newblock In \emph{Proceedings of the 2013 Conference on Empirical Methods in
  Natural Language Processing, {EMNLP} 2013, 18-21 October 2013, Grand Hyatt
  Seattle, Seattle, Washington, USA, {A} meeting of SIGDAT, a Special Interest
  Group of the {ACL}}, pages 1631--1642. {ACL}.

\bibitem[{Talmor et~al.(2019)Talmor, Herzig, Lourie, and Berant}]{csqa}
Alon Talmor, Jonathan Herzig, Nicholas Lourie, and Jonathan Berant. 2019.
\newblock \href {https://doi.org/10.18653/v1/n19-1421} {Commonsenseqa: {A}
  question answering challenge targeting commonsense knowledge}.
\newblock In \emph{Proceedings of the 2019 Conference of the North American
  Chapter of the Association for Computational Linguistics: Human Language
  Technologies, {NAACL-HLT} 2019, Minneapolis, MN, USA, June 2-7, 2019, Volume
  1 (Long and Short Papers)}, pages 4149--4158. Association for Computational
  Linguistics.

\bibitem[{Wang et~al.(2023)Wang, Shang, and Zhong}]{goal_cluster}
Zihan Wang, Jingbo Shang, and Ruiqi Zhong. 2023.
\newblock \href {https://doi.org/10.48550/arXiv.2305.13749} {Goal-driven
  explainable clustering via language descriptions}.
\newblock \emph{CoRR}, abs/2305.13749.

\bibitem[{Wei et~al.(2022{\natexlab{a}})Wei, Bosma, Zhao, Guu, Yu, Lester, Du,
  Dai, and Le}]{wei2022finetuned}
Jason Wei, Maarten Bosma, Vincent Zhao, Kelvin Guu, Adams~Wei Yu, Brian Lester,
  Nan Du, Andrew~M. Dai, and Quoc~V Le. 2022{\natexlab{a}}.
\newblock \href {https://openreview.net/forum?id=gEZrGCozdqR} {Finetuned
  language models are zero-shot learners}.
\newblock In \emph{International Conference on Learning Representations}.

\bibitem[{Wei et~al.(2022{\natexlab{b}})Wei, Tay, Bommasani, Raffel, Zoph,
  Borgeaud, Yogatama, Bosma, Zhou, Metzler et~al.}]{wei2022emergent}
Jason Wei, Yi~Tay, Rishi Bommasani, Colin Raffel, Barret Zoph, Sebastian
  Borgeaud, Dani Yogatama, Maarten Bosma, Denny Zhou, Donald Metzler, et~al.
  2022{\natexlab{b}}.
\newblock Emergent abilities of large language models.
\newblock \emph{arXiv preprint arXiv:2206.07682}.

\bibitem[{Wei et~al.(2022{\natexlab{c}})Wei, Wang, Schuurmans, Bosma, Ichter,
  Xia, Chi, Le, and Zhou}]{cot}
Jason Wei, Xuezhi Wang, Dale Schuurmans, Maarten Bosma, Brian Ichter, Fei Xia,
  Ed~H. Chi, Quoc~V. Le, and Denny Zhou. 2022{\natexlab{c}}.
\newblock \href
  {http://papers.nips.cc/paper\_files/paper/2022/hash/9d5609613524ecf4f15af0f7b31abca4-Abstract-Conference.html}
  {Chain-of-thought prompting elicits reasoning in large language models}.
\newblock In \emph{NeurIPS}.

\bibitem[{Wei and Zou(2019)}]{wei-zou-2019-eda}
Jason Wei and Kai Zou. 2019.
\newblock \href {https://doi.org/10.18653/v1/D19-1670} {{EDA}: Easy data
  augmentation techniques for boosting performance on text classification
  tasks}.
\newblock In \emph{Proceedings of the 2019 Conference on Empirical Methods in
  Natural Language Processing and the 9th International Joint Conference on
  Natural Language Processing (EMNLP-IJCNLP)}, pages 6382--6388, Hong Kong,
  China. Association for Computational Linguistics.

\bibitem[{Whitehouse et~al.(2023)Whitehouse, Choudhury, and
  Aji}]{whitehouse2023llm}
Chenxi Whitehouse, Monojit Choudhury, and Alham~Fikri Aji. 2023.
\newblock Llm-powered data augmentation for enhanced crosslingual performance.
\newblock \emph{arXiv preprint arXiv:2305.14288}.

\bibitem[{Williams et~al.(2018)Williams, Nangia, and Bowman}]{mnli}
Adina Williams, Nikita Nangia, and Samuel~R. Bowman. 2018.
\newblock \href {https://doi.org/10.18653/v1/n18-1101} {A broad-coverage
  challenge corpus for sentence understanding through inference}.
\newblock In \emph{Proceedings of the 2018 Conference of the North American
  Chapter of the Association for Computational Linguistics: Human Language
  Technologies, {NAACL-HLT} 2018, New Orleans, Louisiana, USA, June 1-6, 2018,
  Volume 1 (Long Papers)}, pages 1112--1122. Association for Computational
  Linguistics.

\bibitem[{Xiao et~al.(2018)Xiao, Hong, and Ma}]{elegant}
Taihong Xiao, Jiapeng Hong, and Jinwen Ma. 2018.
\newblock \href {https://doi.org/10.1007/978-3-030-01249-6\_11} {{ELEGANT:}
  exchanging latent encodings with {GAN} for transferring multiple face
  attributes}.
\newblock In \emph{Computer Vision - {ECCV} 2018 - 15th European Conference,
  Munich, Germany, September 8-14, 2018, Proceedings, Part {X}}, volume 11214
  of \emph{Lecture Notes in Computer Science}, pages 172--187. Springer.

\bibitem[{Yang and Klein(2021)}]{fudge}
Kevin Yang and Dan Klein. 2021.
\newblock \href {https://doi.org/10.18653/v1/2021.naacl-main.276} {{FUDGE:}
  controlled text generation with future discriminators}.
\newblock In \emph{Proceedings of the 2021 Conference of the North American
  Chapter of the Association for Computational Linguistics: Human Language
  Technologies, {NAACL-HLT} 2021, Online, June 6-11, 2021}, pages 3511--3535.
  Association for Computational Linguistics.

\bibitem[{Yoo et~al.(2021)Yoo, Park, Kang, Lee, and Park}]{yoo2021gpt3mix}
Kang~Min Yoo, Dongju Park, Jaewook Kang, Sang-Woo Lee, and Woomyeong Park.
  2021.
\newblock Gpt3mix: Leveraging large-scale language models for text
  augmentation.
\newblock \emph{arXiv preprint arXiv:2104.08826}.

\bibitem[{Zhang et~al.(2015)Zhang, Zhao, and LeCun}]{ag}
Xiang Zhang, Junbo~Jake Zhao, and Yann LeCun. 2015.
\newblock \href
  {https://proceedings.neurips.cc/paper/2015/hash/250cf8b51c773f3f8dc8b4be867a9a02-Abstract.html}
  {Character-level convolutional networks for text classification}.
\newblock In \emph{Advances in Neural Information Processing Systems 28: Annual
  Conference on Neural Information Processing Systems 2015, December 7-12,
  2015, Montreal, Quebec, Canada}, pages 649--657.

\bibitem[{Zhou et~al.(2022)Zhou, Zheng, Tang, Jian, and Yang}]{flipda}
Jing Zhou, Yanan Zheng, Jie Tang, Li~Jian, and Zhilin Yang. 2022.
\newblock \href {https://doi.org/10.18653/v1/2022.acl-long.592} {Flipda:
  Effective and robust data augmentation for few-shot learning}.
\newblock In \emph{Proceedings of the 60th Annual Meeting of the Association
  for Computational Linguistics (Volume 1: Long Papers), {ACL} 2022, Dublin,
  Ireland, May 22-27, 2022}, pages 8646--8665. Association for Computational
  Linguistics.

\end{thebibliography}
\bibliographystyle{acl_natbib}

\clearpage

\appendix

\section{Dataset Statistics}
\label{apdx:stat}
\begin{table}[H]
\centering
\scalebox{.9}{
\begin{tabular}{lccc}
\toprule
\small
Dataset & SST-2 & TweetEmo & AG-News \\
\midrule
Domain & Sentiment & Sentiment & Topic \\
\#Test & $1.8K$ & $1.4K$ & $7.6K$ \\
\#Label & $2$ & $4$ & $4$ \\
\midrule
Dataset & MNLI & MRPC & CSQA \\
\midrule
Task & NLI & STS & MCQA \\
\#Test & $9.8K$ & $1.7K$ & $1.1K$ \\
\#Label & $3$ & $2$ & $5$ \\
\midrule
Dataset & $14$RES & $14$LAP & CommonGen \\
\midrule
Task & ABSA & ABSA & NLG \\
\#Test & $0.1K$ & $0.1K$ & $4.0K$ \\
\#Label & $4$ & $4$ & - \\
\bottomrule
\end{tabular}
}
\caption{The statistics of datasets in our experiments.} 
\label{tab:stat}
\end{table}

The statistics of the dataset used in the experiments are presented in Table~\ref{tab:stat}. The numbers of test instances in matched and mismatched are both $9.8K$.

\section{Attribute Names}
\label{apdx:attr}
\begin{table}[H]
\centering
\small
\scalebox{1.}{
\begin{tabular}{ll}
\toprule
Dataset & Attributes \\
\midrule
\multirow{2}*{SST-2} & sentiment: positive \\ 
 & sentiment: negative \\ 
\midrule
\multirow{4}*{TweetEmo} & sentiment: anger \\ 
 & sentiment: joy \\ 
 & sentiment: optimism \\ 
 & sentiment: sadness \\ 
\midrule
\multirow{4}*{AG-News} & topic: world news \\ 
 & topic: sports news \\ 
 & topic: business news \\ 
 & topic: sci/tech news \\ 
\midrule
\multirow{3}*{MNLI} & natural language inference: contradiction \\ 
 & natural language inference: neutral \\ 
 & natural language inference: entailment \\ 
\midrule
\multirow{2}*{MRPC} & semantics: equivalent to sentence 1 \\ 
 & semantics: inequivalent to sentence 1 \\ 
\midrule
\multirow{1}*{CSQA} & best choice: <answer name> \\ 
\bottomrule
\end{tabular}
}
\caption{The attribute names in datasets of our experiments.} 
\label{tab:attr}
\end{table}

The attribute names of the dataset used in the experiments are presented in Table~\ref{tab:attr}.

\newpage

\section{Prompts}
\label{apdx:prompt}
\begin{table}[H]
\centering
\scalebox{1.0}{
\begin{tabular}{p{1.5cm}p{5cm}}
\toprule
Target & Prompt\\
\midrule
CoTAM &  ``<sentence>''\\
& Please think step by step: \\
& 1. What are some other attributes of the above sentence except ``<attr>''?\\
& 2. How to write a similar sentence with these attributes and ``<new attr>''?\\
& 3. Write such a sentence without any other explanation.\\
\midrule
CoTDA &  ``<sentence>''\\
& Please think step by step: \\
& 1. What are some other attributes of the above sentence except ``<attr>''?\\
& 2. How to write a similar sentence with these attributes and ``<attr>''?\\
& 3. Write such a sentence without any other explanation.\\
\midrule
FlipDA &  ``<sentence>''\\
& Please think step by step: \\
& 1. How to switch the above sentence to ``<new attr>'' by changing some spans?\\
& 2. Write the switched sentence without any other explanation.\\
\bottomrule
\end{tabular}
}
\caption{The prompts used in our experiments.} 
\label{tab:prompt}
\end{table}

The prompts used in the experiments are presented in Table~\ref{tab:prompt}.

\newpage

\section{Results on Open-sourced LLMs}
\label{apdx:open}
\begin{table*}
\centering
\small
\begin{tabular}{llccccccc}
\toprule
\multicolumn{2}{l}{\textbf{Method}} & SST-2 & TweetEmo & AG-NEWS & MNLI$_{\textrm{m}}$ & MNLI$_{\textrm{mm}}$ & MRPC & CSQA \\
\midrule
\multirow{6}*{\rotatebox{90}{Fine-tuning}} & Base & $60.54$ & $44.38$ & $81.05$ & $35.88$ & $38.75$ & $51.96$ & $34.54$ \\
& Extra Annotation$^\dag$ & $62.17$ & $69.51$ & $88.66$ & $43.33$ & $44.03$ & $57.50$ & $47.36$ \\
& LLM Pseudo Label & $61.37$ & $66.85$ & $78.51$ & $36.55$ & $\textbf{39.48}$ & $56.87$ & $40.65$ \\
& FlipDA++ & $73.91$ & $66.94$ & $82.56$ & $36.92$ & $37.98$ & $61.03$ & $38.16$ \\
& CoTDA & $71.77$ & $63.46$ & $83.03$ & $35.40$ & $36.02$ & $51.71$ & $36.75$ \\
& \method & $\textbf{74.83}$ & $\textbf{67.97}$ & $\textbf{83.49}$ & $\textbf{37.21}$ & $38.52$ & $\textbf{61.96}$ & $\textbf{42.88}$ \\
\midrule
\multirow{6}*{\rotatebox{90}{ICL}} & Base & $88.40$ & $65.20$ & $44.00$ & $67.60$ & $67.00$ & $74.20$ & $68.10$ \\
& Extra Annotation$^\dag$ & $88.90$ & $77.80$ & $76.20$ & $69.10$ & $68.00$ & $76.90$ & $74.20$ \\
& LLM Pseudo Label & $88.60$ & $\textbf{74.20}$ & $56.90$ & $\textbf{68.00}$ & $66.50$ & $75.10$ & $68.90$ \\
& FlipDA++ & $\textbf{89.90}$ & $71.80$ & $68.30$ & $63.60$ & $64.50$ & $\textbf{75.80}$ & $70.20$ \\
& CoTDA & $88.70$ & $69.90$ & $73.90$ & $67.60$ & $66.50$ & $68.70$ & $68.40$ \\
& \method & $\textbf{89.90}$ & $72.80$ & $\textbf{75.30}$ & $67.90$ & $\textbf{67.20}$ & $74.60$ & $\textbf{70.80}$ \\
\bottomrule
\end{tabular}
\caption{Repeated experiments on Table~\ref{tab:main} with open-sourced LLM (\texttt{llama-2-70b-chat-hf}), which is applied to both data generation and in-context learning.}
\label{tab:main-open}
\end{table*}

To improve the reproducibility of our results, we present results on open-sourced LLMs as presented in Table~\ref{tab:main-open}. The performance of different augmentation methods on the open-sourced LLM is generally consistent with our main experiments. 

\section{Attribute Statistics}
\label{apdx:stats}
\begin{figure}[H]
\centering
\includegraphics[width=0.49\textwidth]{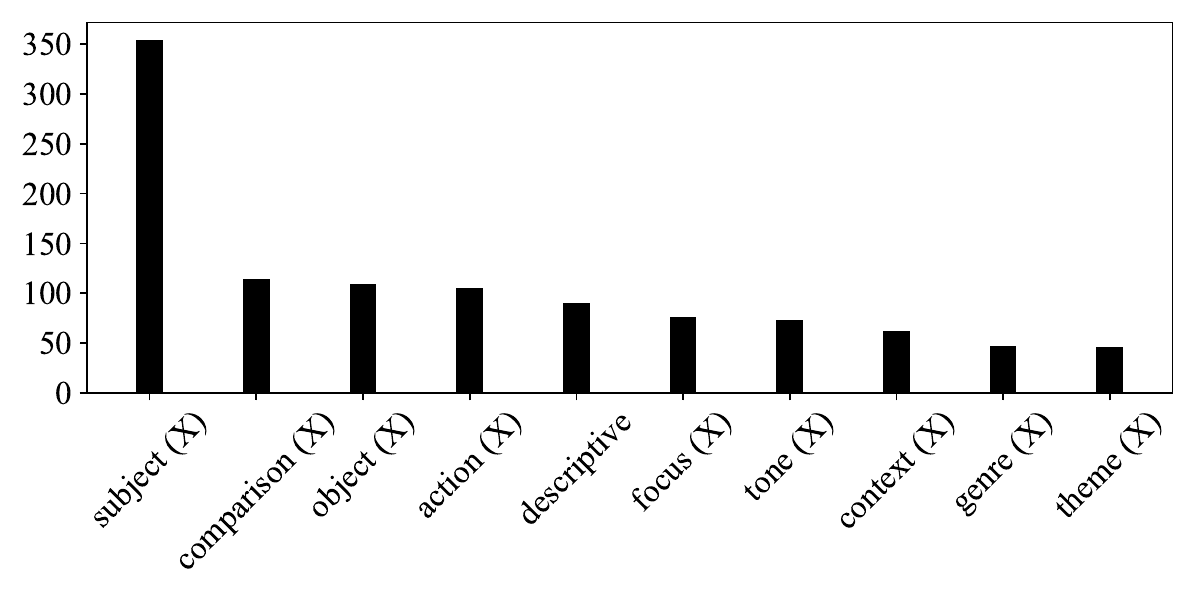}
\caption{The statistics the most frequent $10$ attributes in the decomposition step of \method.}
\label{fig:attribute_frequency}
\end{figure}

In this section, we further explore the dynamic attribute decomposition mechanism in \method. For $1315$ instances from SST-2, there are $4513$ decomposed attributes ($3.43$ per instance) and $2409$ different ones. The distribution is in a long-tail pattern with $2124$ attributes only appearing once. We show the statistics the most frequent $10$ attributes from the decomposition in Table~\ref{fig:attribute_frequency}. We can observe a semantic diversity among the attributes, which verifies the ability of LLMs to comprehend the features of different inputs. As the most popular attribute \textit{subject (X)} only appear in about $20\%$, there is no dominant attribute in the decomposition, which shows the flexibility of LLM-driven feature analysis. We also provide a quantitative comparison with a fixed feature pool in the ablation study. 

\begin{figure}[H]
\centering
\includegraphics[width=0.49\textwidth]{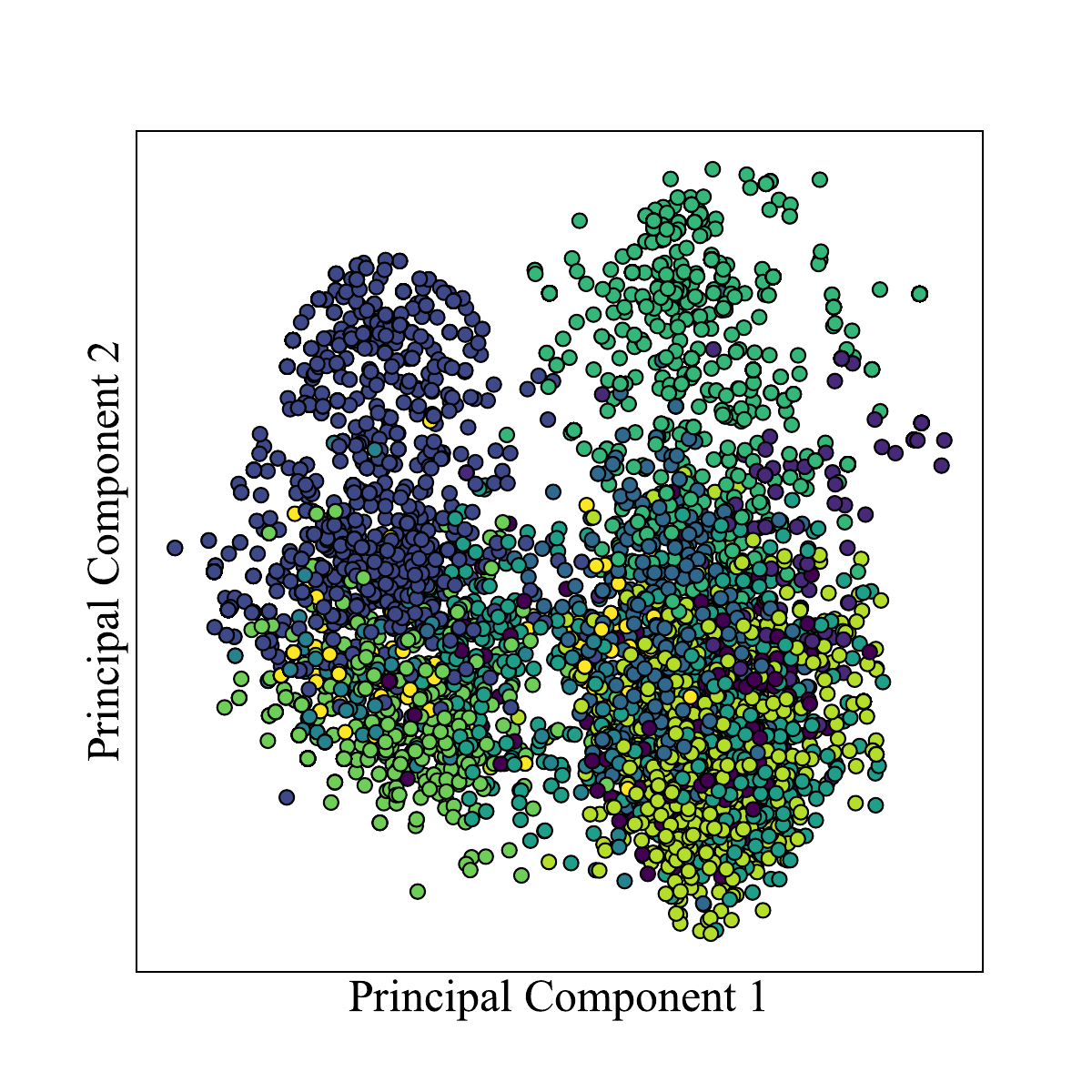}
\caption{The statistics the most frequent $10$ attributes in the decomposition step of \method.}
\label{fig:attribute_pca}
\end{figure}

\paragraph{Clustering Analysis} We also present an analysis of the semantic distribution of the attributes. We use the sentence encoder to encode the attribute names and run K-Means ($K=10$) clustering, which is presented in Figure~\ref{fig:attribute_pca}. We find the attributes that are the closest to the cluster centers and get $10$ attributes: 

\begin{itemize}
    \item an implied evaluation or opinion
    \item a passive tone
    \item subject (X)
    \item subjectivity
    \item action (X)
    \item it mentions specific elements of a film (X)
    \item brevity: short phrase
    \item focus (X)
    \item evocative
    \item comparison (X)
\end{itemize}

\noindent which shows the diversity of the discovered attributes and improves our understanding of the attribute distribution in the pipeline.

\newpage

\section{Pipeline Comparison}

We use a simple case study to compare different text augmentation methods. We can see CoTDA is not very different from conventional data augmentation. While FlipDA++ successfully flips the selected attribute, it fails to preserve the attribute "contrast: between actors' performance and overall movie experience". The data generated from our proposed CoTAM best satisfies our demand to switch selected attributes and keep others constant.

\begin{table*}
\centering
\scalebox{1.0}{
\begin{tabular}{p{1.5cm}p{13.5cm}}
\toprule
Text & While some actors perform well, the movie is dull overall.\\
Attribute & ``Sentiment: Negative'' $\rightarrow$ ``Sentiment: Positive''\\
\midrule
CoTAM &  1. Other attributes of the sentence include: "subject: movie", "aspect: actors' performance", "aspect: overall movie experience", "contrast: between actors' performance and overall movie experience".\\
& 2. To write a similar sentence with these attributes and "sentiment: positive", we need to maintain the subject and aspects, but change the sentiment. We can also keep the contrast, but reverse it.\\
& 3. \textbf{"While some scenes are lackluster, the movie is captivating overall."}\\
\midrule
CoTDA & 1. Other attributes of the sentence include: comparison (comparing the performance of the actors to the overall movie), subjectivity (the speaker's personal opinion about the movie), and contrast (between the actors' performance and the overall movie).\\
& 2. To write a similar sentence with these attributes and a negative sentiment, we need to maintain the comparison, subjectivity, and contrast. We could talk about another form of entertainment, such as a book or a music album, and compare different aspects of it.\\
& 3. \textbf{"While the lyrics are thought-provoking, the album fails to impress overall."}\\
\midrule
FlipDA &  1. To switch the sentiment of the sentence to positive, we need to change the negative parts of the sentence. The negative part in this sentence is "the movie is dull overall". We can change this to "the movie is exciting overall".\\
& 2. \textbf{"While some actors perform well, the movie is exciting overall."}\\
\bottomrule
\end{tabular}
}
\caption{The prompts used in our experiments.} 
\label{tab:prompt}
\end{table*}

\end{document}